\pdfoutput=1

\documentclass[11pt]{article}

\usepackage[]{EMNLP2022}

\usepackage{times}
\usepackage{latexsym}

\usepackage[T1]{fontenc}

\usepackage[utf8]{inputenc}

\usepackage{microtype}

\usepackage{inconsolata}

\usepackage{multirow}
\usepackage{graphicx}
\usepackage{booktabs}
\usepackage{subcaption}
\usepackage{tabu}
\usepackage{amsmath}
\usepackage{verbatim}

\newcommand{\interalia}[1]{\citep[\emph{inter alia}]{#1}}

\title{NarrowBERT: Accelerating Masked Language Model \\Pretraining and Inference}

\author{Haoxin Li$^1$ \quad Phillip Keung$^3$ \quad Daniel Cheng$^1$ \quad Jungo Kasai$^1$ \quad Noah A. Smith$^{1,2}$ \\
  $^1$Paul G. Allen School of Computer Science \& Engineering, University of Washington, USA \\
  $^2$Allen Institute for Artificial Intelligence, USA \\
  $^3$Department of Statistics, University of Washington, USA \\
  \texttt{\{lihaoxin,d0,jkasai,nasmith\}@cs.washington.edu}, 
  \texttt{pkeung@uw.edu} \\
  }

\begin{document}
\maketitle
\begin{abstract}
Large-scale language model pretraining is a very successful form of self-supervised learning in natural language processing, but it is increasingly expensive to perform as the models and pretraining corpora have become larger over time. We propose NarrowBERT, a modified transformer encoder that increases the throughput for masked language model pretraining by more than $2\times$. NarrowBERT sparsifies the transformer model such that the self-attention queries and feedforward layers only operate on the masked tokens of each sentence during pretraining, rather than all of the tokens as with the usual transformer encoder. We also show that NarrowBERT increases the throughput at inference time by as much as $3.5\times$ with minimal (or no) performance degradation on sentence encoding tasks like MNLI. Finally, we examine the performance of NarrowBERT on the IMDB and Amazon reviews classification and CoNLL NER tasks and show that it is also comparable to standard BERT performance.
\end{abstract}

\section{Introduction}
Pretrained masked language models, such as BERT \cite{devlins2019bert}, RoBERTa \cite{Liu2019RoBERTaAR}, and DeBERTa \cite{deberta}, have pushed the state-of-the-art on a wide range of downstream tasks in natural language processing.
At their core is the transformer architecture \cite{Vaswani2017AttentionIA} that consists of interleaved self-attention and feedforward sublayers.
Since the former sublayer implies quadratic time complexity in the input sequence length \cite{Vaswani2017AttentionIA}, many have proposed methods to make the self-attention computation more efficient \interalia{katharopoulos-et-al-2020,performer,wang2020linformer,RFA,peng-etal-2022-abc}.

In this work, we explore an orthogonal approach to efficiency: can we make masked language models efficient by \textit{reducing} the length of the input sequence that each layer needs to process?
In particular, pretraining by masked language modeling only involves prediction of masked tokens (typically, only 15\% of the input tokens; \citealp{devlins2019bert,Liu2019RoBERTaAR}).
Despite this sparse pretraining objective, each transformer layer computes a representation for every token.
In addition to pretraining, many downstream applications only use a single vector representation (i.e., only the \texttt{[CLS]} token) for prediction purposes, which is much smaller than the number of input tokens (e.g., sequence classification tasks as in GLUE/SuperGLUE; \citealp{wang-etal-2018-glue,superglue}).
By narrowing the input sequence for transformer layers, we can accelerate both pretraining and inference.

\begin{figure*}[ht]
\begin{subfigure}{1\textwidth}
  \centering
  \includegraphics[width=12cm]{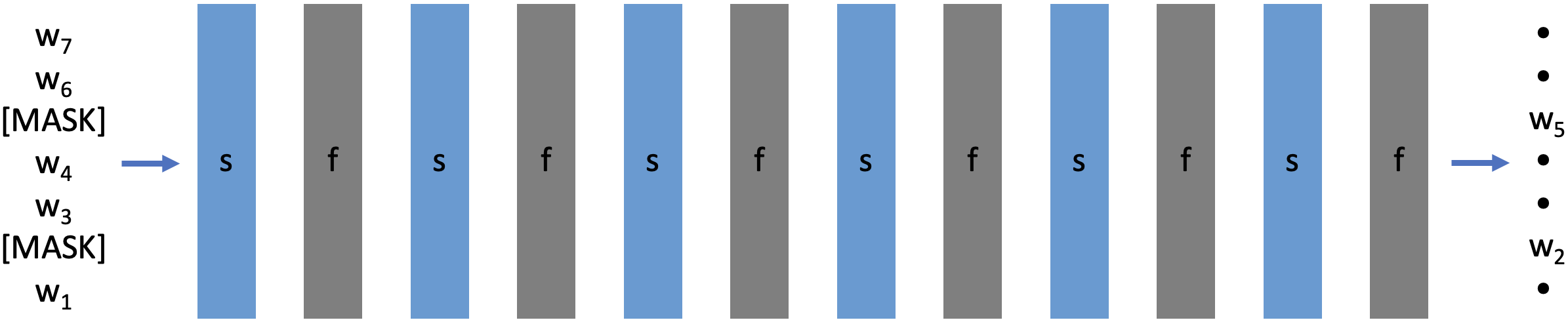}
  \caption{\texttt{\{6,sf\}} model: standard BERT with the transformer encoder, trained on MLM loss.}
  \label{fig:normal}
\end{subfigure}
\\ 
\begin{subfigure}{1\textwidth}
	\centering
	\includegraphics[width=12cm]{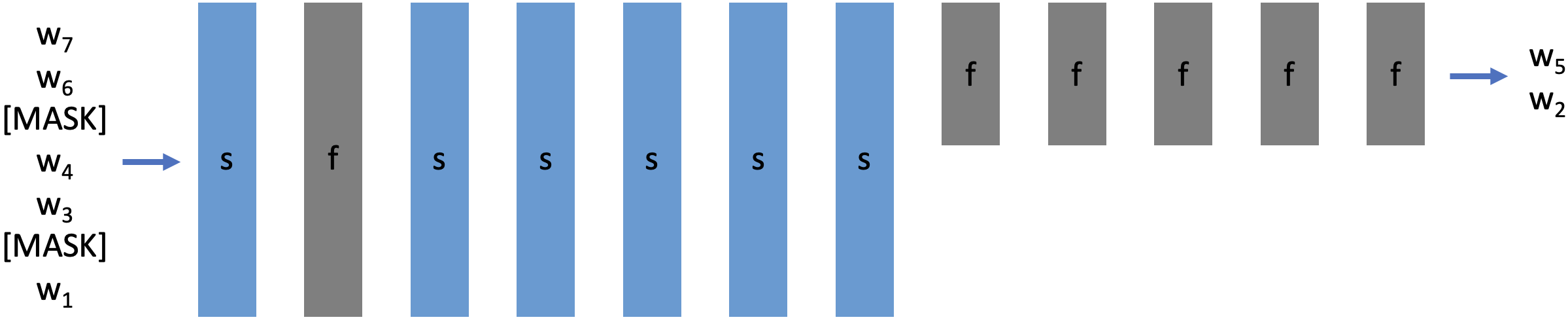}
	\caption{\texttt{sf\{5,s\}:\{5,f\}} ContextFirst model: Transformer encoder with re-ordered layers. Attentional contextualization is performed all-at-once near the beginning of the model.}
	\label{fig:context_first}
\end{subfigure}
\\ \qquad
\begin{subfigure}{1\textwidth}
  \centering
  \includegraphics[width=12cm]{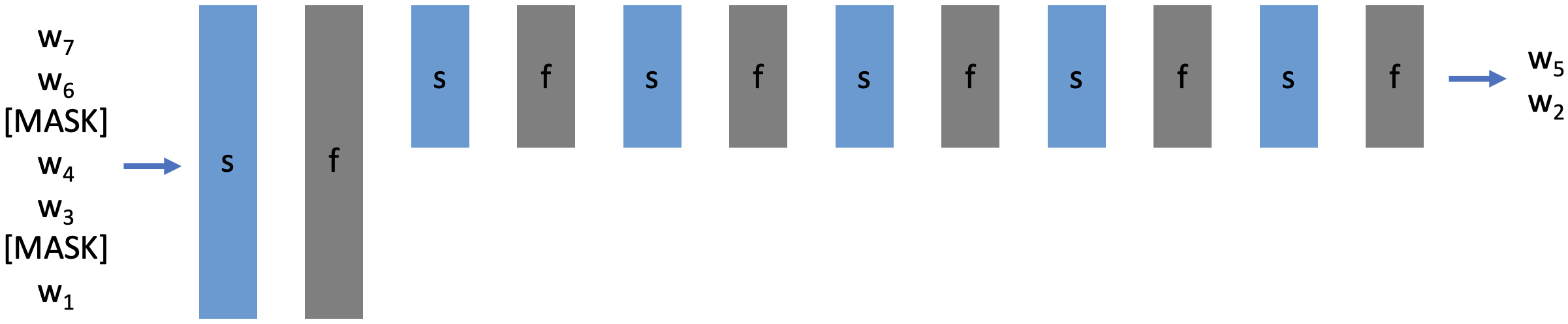}
  \caption{\texttt{sf:\{5,sf\}} SparseQueries model: Transformer encoder with sparsified queries. Contextualization is focused on [MASK] tokens only. (See Fig. \ref{fig:narrow_queries}.)}
  \label{fig:narrow_first}
\end{subfigure}
\caption{Examples of standard BERT and NarrowBERT variations. NarrowBERT takes advantage of the sparsity in the masking (i.e., only 15\% of tokens need to be predicted) to reduce the amount of computation in the transformer encoder.}
\label{fig:narrowbert}
\end{figure*}

We present NarrowBERT, a new architecture that takes advantage of the sparsity in the training objective. We present two NarrowBERT methods in the sections that follow (Figure \ref{fig:narrowbert}). We provide the code to reproduce our experiments at \url{https://github.com/lihaoxin2020/narrowbert}.
The first method reduces the input sequence for the feedforward sublayers by reordering the interleaved self-attention and feedforward sublayers in the standard transformer architecture \cite{press-etal-2020-improving}: after two standard, interleaved transformer layers, self-attention sublayers are first applied, followed only by feedforward sublayers.
This way, the feedforward sublayer computations are only performed for \emph{masked tokens}, resulting in a $1.3\times$ speedup in pretraining (\S\ref{section:experiments}).
The second approach reduces the input length to the attention sublayers: \textit{queries} are only computed for masked tokens in the attention mechanism \cite{Bahdanau2014NeuralMT}, while the \textit{keys} and \textit{values} are not recomputed for non-masked tokens, which leads to a greater than $2\times$ speedup in pretraining.

We extensively evaluate our efficient pretrained models on well-established downstream tasks (e.g., \citealp{wang-etal-2018-glue, tjong-kim-sang-de-meulder-2003-introduction}.)
We find that our modifications result in almost no drop in downstream performance, while providing substantial pretraining and inference speedups (\S\ref{section:experiments}).
While efficient attention variants are a promising research direction, this work presents a different and simple approach to making transformers efficient, with minimal changes in architecture.


\section{NarrowBERT}

\begin{figure*}[ht]
  \centering
  \includegraphics[width=12cm]{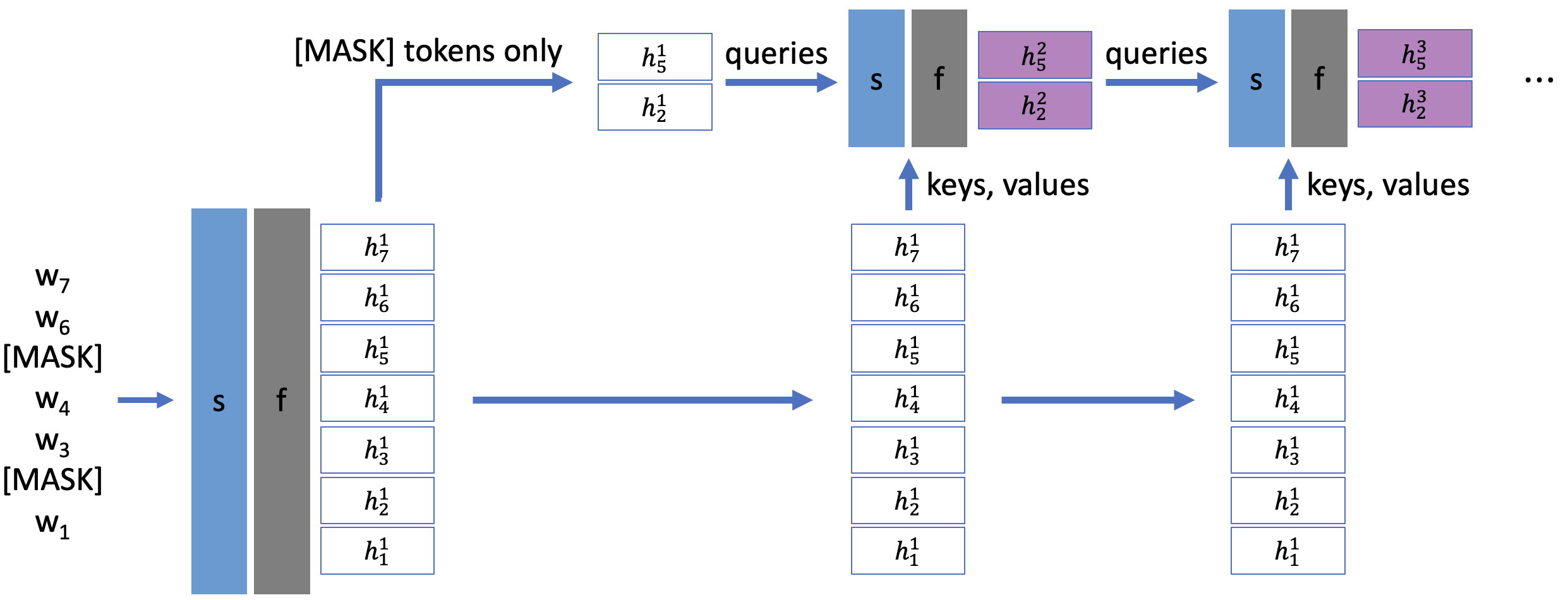}
\caption{Sparse queries in the attention layers. Only the masked positions are contextualized as query vectors in subsequent \texttt{s} layers. The inputs are contextualized once by the first \texttt{s} layer and \texttt{f} layer, and reused as the keys and values in all subsequent attention layers.}
\label{fig:narrow_queries}
\end{figure*}
\begin{table*}[h]
	\small
	\centering
\begin{tabular}{@{}lccccccccc@{}}
	\toprule
                   & Pretrain &   Finetune           &           Inference        & \multicolumn{6}{c}{\textbf{GLUE}}                   \\
                   & Speedup &  Speedup &  Speedup & MNLI & QNLI & SST2 & STS-B & QQP  & WNLI   \\
                   \midrule
Baseline BERT  (\texttt{\{12,sf\}})    & $1\times$ & $1\times$          & $1\times$      & 0.83 & 0.91 & 0.93 & 0.89  & 0.87 & 0.56    \\
Funnel Transformer (B4-4-4) & 0.88$\times$ & 0.86$\times$ & 0.78$\times$  & 0.78  & 0.87 & 0.88 & 0.86 & 0.86 & 0.56  \\
ContextFirst & 1.33$\times$  & 1.24$\times$           & 1.64$\times$               & 0.82 & 0.90  & 0.91 & 0.89 & 0.87 & 0.56     \\
SparseQueries: &&& &&& && \\
\quad \texttt{\{1,sf\}:\{11,sf\}} & 2.47$\times$ & $4.73\times$              & 4.64$\times$               & 0.77 & 0.87 & 0.89 & 0.84 & 0.80 & 0.56       \\
\quad \texttt{\{2,sf\}:\{10,sf\}} & 2.34$\times$ & 2.82$\times$               & 3.49$\times$               & 0.81 & 0.88 & 0.91 & 0.88 & 0.87 & 0.59     \\
\quad \texttt{\{3,sf\}:\{9,sf\}} & 2.15$\times$ & 2.43$\times$ & 2.79$\times$ & 0.81 & 0.89 & 0.91 & 0.86 & 0.87 & 0.56      \\
\quad \texttt{\{4,sf\}:\{8,sf\}} & 1.63$\times$ & 2.13$\times$ & 2.33$\times$ & 0.82 & 0.88 & 0.91 & 0.89 & 0.87 & 0.57     \\
\bottomrule
\end{tabular}
\caption{Test scores on various GLUE tasks. (`MNLI' scores refer to the MNLI matched dev set.) Finetuning and inference speedups refer to speeds on the MNLI task. ContextFirst is equivalent to \texttt{sfsf\{10,s\}:\{10,f\}} in our model notation.}
\label{table:extrinsic}
\end{table*}

In Figures \ref{fig:context_first} and \ref{fig:narrow_first}, we illustrate two variations of NarrowBERT. We define some notation to describe the configuration of our models. \texttt{s} refers to a \textbf{single self-attention layer} and \texttt{f} refers to a \textbf{single feedforward layer}. The colon \texttt{:} refers to the \textbf{`narrowing' operation}, which gathers the masked positions from the output of the previous layer.

The first variation (`ContextFirst' in Fig.~\ref{fig:context_first}) uses attention to contextualize all-at-once at the beginning of the model. In short, the transformer layers have been rearranged to frontload the attention components. The example given in the figure specifies the model as \texttt{sf\{5,s\}:\{5,f\}}, which means that the input sentence is encoded by a self-attention layer, a feedforward layer, and 5 consecutive self-attention layers. At that point, the masked positions from the encoded sentence are gathered into a tensor and passed through 5 feedforward layers, \textbf{thereby avoiding further computations for all unmasked tokens}.
Finally, the masked positions are unmasked and the MLM loss is computed.

The second variation (`SparseQueries' in Fig.~\ref{fig:narrow_first}) does not reorder the layers at all. Instead, the \texttt{sf:\{5,sf\}} model contextualizes the input sentence in a more limited way. As shown in Figure \ref{fig:narrow_queries}, the input sentence is first contextualized by a \texttt{s} and a \texttt{f} layer, but the non-masked tokens are never contextualized again afterwards. Only the masked tokens are contextualized by the remaining \texttt{\{5,sf\}} layers.

Since the masked tokens are only about 15\% of the total sentence length, the potential speedup is \char`\~$6.6\times$ for every feedforward or attention layer downstream of a narrowing \texttt{:} operation. The memory usage can also decrease by  \char`\~$6.6\times$ for those layers since the sequence length has decreased, which allows us to use larger batch sizes during training.

For GLUE, Amazon, and IMDB text classification tasks, only the \texttt{[CLS]} token is used for prediction. When we finetune or predict with ContextFirst on a GLUE/Amazon/IMDB task, the feedforward layers only need to operate on the \texttt{[CLS]} token. When we finetune or predict with SparseQueries, only the \texttt{[CLS]} token is used in the queries of the attention layers. Everything after the narrowing \texttt{:} operation only operates on the [CLS] token, which dramatically speeds up the NarrowBERT variants.

\section{Experiments}
\label{section:experiments}

\begin{figure*}[ht]

\begin{subfigure}{.5\textwidth}
  \centering
  \includegraphics[width=7cm]{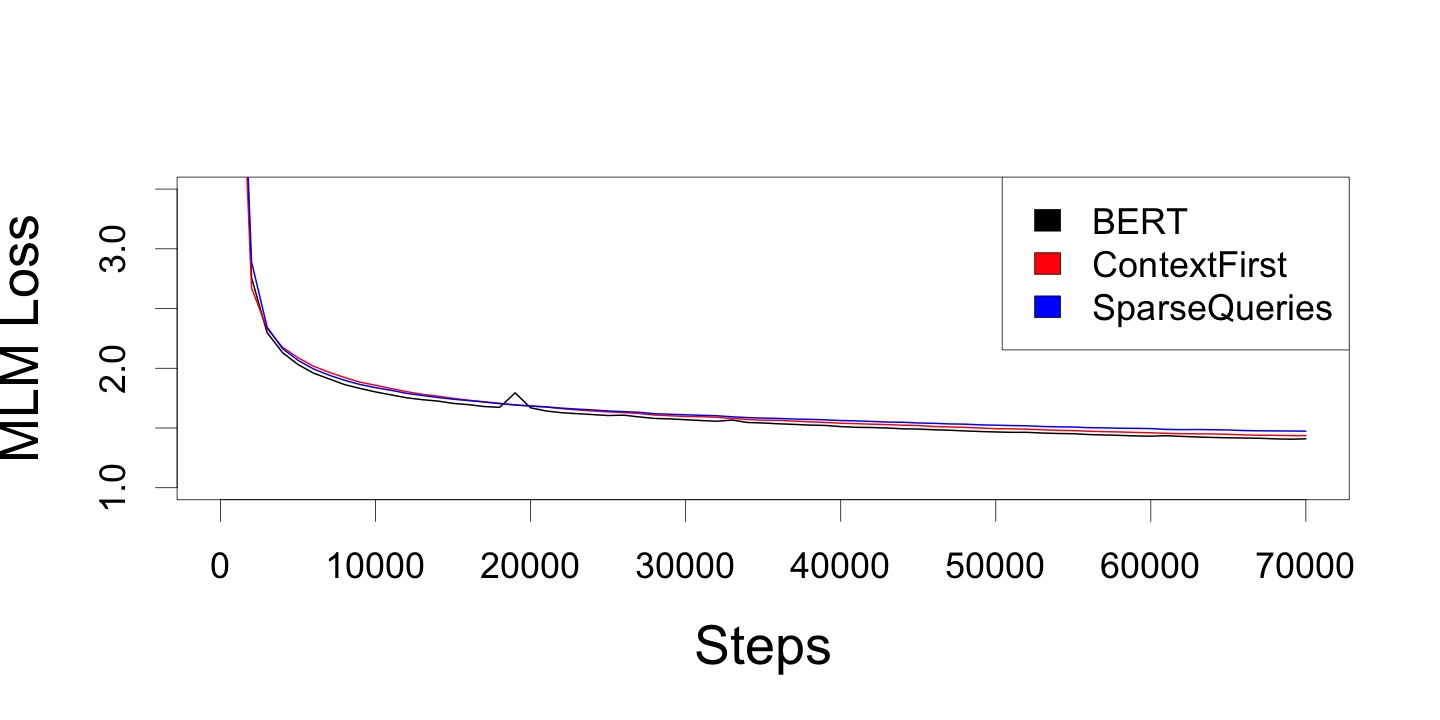}
  \caption{All training steps.}
\end{subfigure}
\qquad
\begin{subfigure}{.5\textwidth}
	\centering
	\includegraphics[width=7cm]{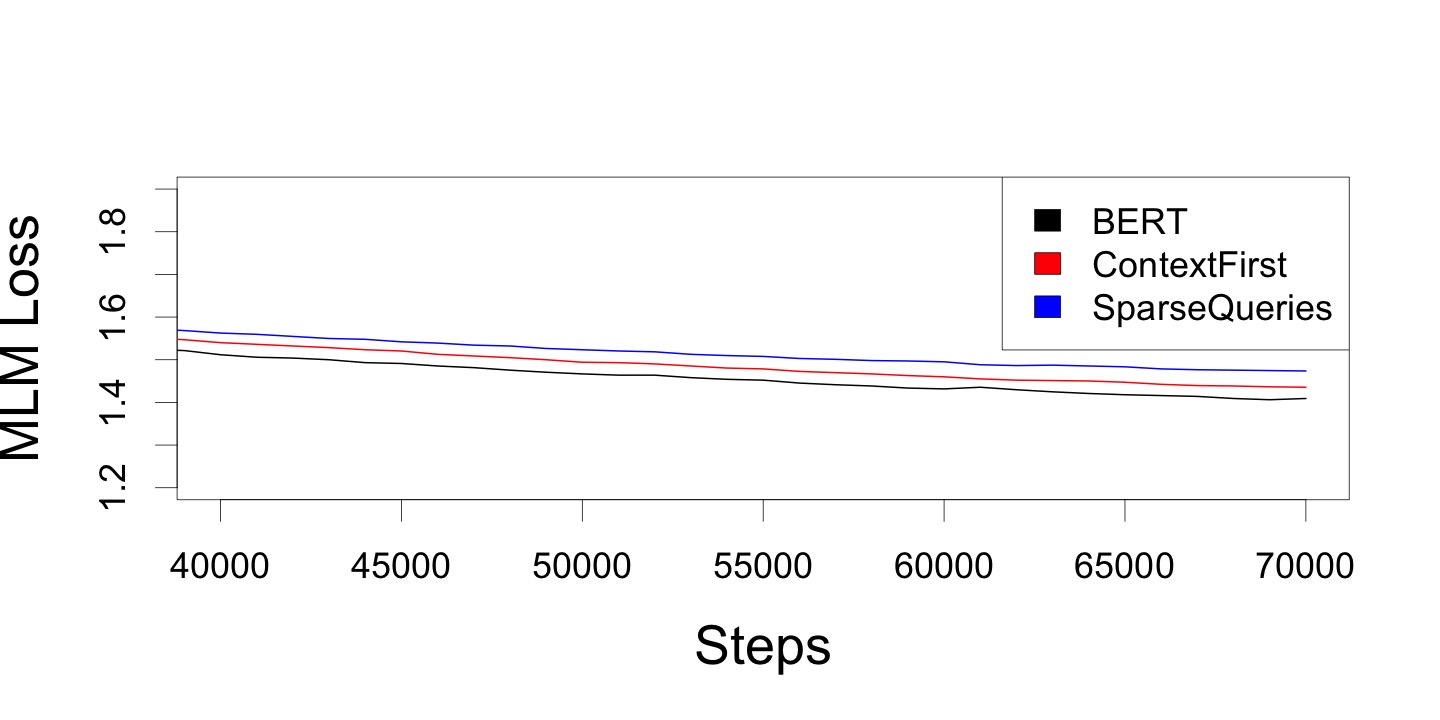}
	\caption{Near the end of training.}
\end{subfigure}
\caption{Development MLM loss over the course of pretraining. At the end of training, the BERT, ContextFirst, and SparseQueries (\texttt{\{2,sf\}:\{10,sf\}}) dev.~MLM losses are 1.41, 1.43, and 1.47 respectively.}
\label{fig:mlm_loss}
\end{figure*}

We focus on 2 models in our experiments: ContextFirst (\texttt{sfsf\{10,s\}:\{10,f\}}) and SparseQueries (\texttt{\{1,sf\}:\{11,sf\}}, $\cdots$ , \texttt{\{4,sf\}:\{8,sf\}}). Our NarrowBERT models all contain 12 self-attention and 12 feedforward layers in total, with the narrowing operation used at different points in the model. We compare NarrowBERT with the baseline BERT model and the Funnel Transformer model \citep{funnel2020}, which is a pretrained encoder-decoder transformer model where the encoder goes through a sequence of length bottlenecks.

In our experiments, we use 15\% masking in masked language model (MLM) training. Following \citet{Liu2019RoBERTaAR}, we do not use next sentence prediction as a pretraining task. We use large batch sizes and high learning rates to fully utilize GPU memory, as suggested in \citet{izsak-etal-2021-train}. Batches are sized to be the largest that fit in GPU memory. We use a learning rate of 0.0005. Models are trained for 70k steps, where each step contains 1728 sequences of 512 tokens, and gradient accumulation is used to accumulate the minibatches needed per step. Models were trained on hosts with 8 Nvidia A100 GPUs. We used the Hugging Face implementations of the baseline BERT and Funnel Transformer models. We pretrained the baseline BERT, Funnel Transformer, and NarrowBERT models using the same Wikipedia and Books corpora and total number of steps. 

\begin{table*}[h]
	\small
	\centering
\begin{tabular}{@{}lcccc@{}}
	\toprule
                   & CoNLL NER & IMDB & Amazon2 & Amazon5 \\
                   \midrule
Baseline BERT  (\texttt{\{12,sf\}})    & 0.90 & 0.93 & 0.96 & 0.66 \\
Funnel Transformer  & 0.87 & 0.92 & 0.95 & 0.65 \\
ContextFirst (\texttt{sfsf\{10,s\}:\{10,f\}}) & 0.89 & 0.93 & 0.95 & 0.65 \\
SparseQueries: && && \\
\quad \texttt{\{1,sf\}:\{11,sf\}} & 0.87 & 0.91 & 0.94 & 0.65 \\
\quad \texttt{\{2,sf\}:\{10,sf\}} & 0.89 & 0.91 & 0.95 & 0.65 \\
\quad \texttt{\{3,sf\}:\{9,sf\}} & 0.89 & 0.92 & 0.95 & 0.65 \\
\quad \texttt{\{4,sf\}:\{8,sf\}} & 0.89 & 0.93 & 0.95 & 0.65 \\
\bottomrule
\end{tabular}
\caption{Test scores on CoNLL NER, IMDB, binarized Amazon reviews, and 5-star Amazon reviews tasks.}
\label{table:nonglue}
\end{table*}

In Figure \ref{fig:mlm_loss}, we see the evolution of the development MLM loss over the course of model training. The BERT and NarrowBERT models all converge to similar values, with the NarrowBERT models reaching a slightly higher MLM loss near the end of training. 

We report the accuracy for MNLI \cite{mnli-dataset}, QNLI \cite{rajpurkar-etal-2016-squad}, SST2 \cite{sst2-dataset}, WNLI \cite{wnli}, IMDB \cite{imdb}, and English Amazon reviews \cite{amazon-reviews}, F1 for QQP \cite{qqp} and CoNLL-2003 NER \cite{tjong-kim-sang-de-meulder-2003-introduction}, and Spearman correlation for STS-B \cite{sts-b}. For the Amazon reviews corpus, we consider both the usual 5-star prediction task and the binarized (i.e., 1--2 stars versus 4--5 stars) task.

In Table \ref{table:extrinsic}, we present the results for our extrinsic evaluation on various GLUE tasks. The reduction in performance is small or non-existent, and on WNLI, the NarrowBERT variations perform better than the baseline. For SparseQueries, it is clear that using more layers prior to the narrowing operation improves performance, though the training and inference speedups become smaller. We note that the Funnel Transformer implementation in Pytorch is slower than the baseline BERT model; this may be due to the fact that the original implementation was written in Tensorflow and optimized for Google TPUs.\footnote{\citet{funnel2020} claim to achieve finetuning FLOPs $0.58\times$ the BERT baseline's. See \url{https://github.com/laiguokun/Funnel-Transformer}.}


It is well known that the variability in the performance of BERT on certain GLUE tasks is extreme \citep{glue_instability, finetuning_tricks, mixout}, where the differences in performance between finetuning runs can exceed $20\%$ (absolute). We have also observed this extreme variability in the course of our own GLUE finetuning experiments. While many techniques have been proposed to address this issue, it is not the goal of this work to apply finetuning stabilization methods to maximize BERT's performance. For this reason, we have excluded the RTE, MRPC, and COLA tasks (which are high-variance tasks studied in the aforementioned papers) from our evaluation.

In Table \ref{table:nonglue}, we provide results on the IMDB and Amazon reviews classification tasks and the CoNLL NER task. Generally, NarrowBERT is close to the baseline in performance, and the SparseQueries performance improves as more layers are used before the narrowing operation.
\section{Discussion and Conclusion}

We have explored two straightforward ways of exploiting the sparsity in the masked language model loss computations: rearranging the layers of the transformer encoder to allow the feedforward components to avoid computations on the non-masked positions, and sparsifying the queries in the attention mechanism to only contextualize the masked positions. The NarrowBERT variants can speed up training by a factor of \char`\~$2\times$ and inference by a factor of \char`\~$3\times$, while maintaining very similar performance on GLUE, IMDB, Amazon, and CoNLL NER tasks. Based on the favorable trade-off between speed and performance seen in Section \ref{section:experiments}, we recommend that practitioners consider using the SparseQueries NarrowBERT model with 2 or 3 layers before narrowing.

\section*{Limitations}
Due to our budget constraint, we only performed pretraining and downstream experiments with base-sized transformer models.
We also only applied the masked language modeling objective, but there are other effective pretraining objectives  (e.g., \citealp{clark2020electra}).
Nonetheless, since we introduced minimal changes in architecture, we hope that subsequent work will benefit from our narrowing operations and conduct a wider range of pretraining and downstream experiments.
While pretrained models can be applied to even more downstream tasks, we designed a reasonable task suite in this work, consisting of both GLUE sentence classification and the CoNLL NER sequential classification tasks.

\section*{Acknowledgments}

The authors thank the anonymous reviewers and Ofir Press at the University of Washington for helpful feedback.  This research was supported in part by NSF grant 2113530.

\bibliography{emnlp2022}

\begin{thebibliography}{28}
\expandafter\ifx\csname natexlab\endcsname\relax\def\natexlab#1{#1}\fi

\bibitem[{Bahdanau et~al.(2015)Bahdanau, Cho, and
  Bengio}]{Bahdanau2014NeuralMT}
Dzmitry Bahdanau, Kyunghyun Cho, and Yoshua Bengio. 2015.
\newblock \href {https://arxiv.org/abs/1409.0473} {Neural machine translation
  by jointly learning to align and translate}.
\newblock In \emph{Proc. of ICLR}.

\bibitem[{Cer et~al.(2017)Cer, Diab, Agirre, Lopez-Gazpio, and Specia}]{sts-b}
Daniel Cer, Mona Diab, Eneko Agirre, Inigo Lopez-Gazpio, and Lucia Specia.
  2017.
\newblock Semeval-2017 task 1: Semantic textual similarity-multilingual and
  cross-lingual focused evaluation.
\newblock \emph{arXiv preprint arXiv:1708.00055}.

\bibitem[{Choromanski et~al.(2021)Choromanski, Likhosherstov, Dohan, Song,
  Gane, Sarl{\'{o}}s, Hawkins, Davis, Mohiuddin, Kaiser, Belanger, Colwell, and
  Weller}]{performer}
Krzysztof Choromanski, Valerii Likhosherstov, David Dohan, Xingyou Song,
  Andreea Gane, Tam{\'{a}}s Sarl{\'{o}}s, Peter Hawkins, Jared Davis, Afroz
  Mohiuddin, Lukasz Kaiser, David Belanger, Lucy Colwell, and Adrian Weller.
  2021.
\newblock \href {https://arxiv.org/abs/2009.14794} {Rethinking attention with
  {Performers}}.
\newblock In \emph{Proc.\ of ICLR}.

\bibitem[{Clark et~al.(2020)Clark, Luong, Le, and Manning}]{clark2020electra}
Kevin Clark, Minh-Thang Luong, Quoc~V. Le, and Christopher~D. Manning. 2020.
\newblock \href {https://openreview.net/pdf?id=r1xMH1BtvB} {{ELECTRA}:
  Pre-training text encoders as discriminators rather than generators}.
\newblock In \emph{Proc.\ of ICLR}.

\bibitem[{Dai et~al.(2020)Dai, Lai, Yang, and Le}]{funnel2020}
Zihang Dai, Guokun Lai, Yiming Yang, and Quoc Le. 2020.
\newblock \href {https://arxiv.org/abs/2006.03236} {Funnel-transformer:
  Filtering out sequential redundancy for efficient language processing}.
\newblock In \emph{Proc.\ of NeurIPS}.

\bibitem[{Devlin et~al.(2019)Devlin, Chang, Lee, and
  Toutanova}]{devlins2019bert}
Jacob Devlin, Ming-Wei Chang, Kenton Lee, and Kristina Toutanova. 2019.
\newblock \href {https://arxiv.org/abs/810.04805} {{BERT}: Pre-training of deep
  bidirectional transformers for language understanding}.
\newblock In \emph{Proc. of NAACL}.

\bibitem[{Dodge et~al.(2020)Dodge, Ilharco, Schwartz, Farhadi, Hajishirzi, and
  Smith}]{finetuning_tricks}
Jesse Dodge, Gabriel Ilharco, Roy Schwartz, Ali Farhadi, Hannaneh Hajishirzi,
  and Noah Smith. 2020.
\newblock Fine-tuning pretrained language models: Weight initializations, data
  orders, and early stopping.
\newblock \emph{arXiv preprint arXiv:2002.06305}.

\bibitem[{He et~al.(2021)He, Liu, Gao, and Chen}]{deberta}
Pengcheng He, Xiaodong Liu, Jianfeng Gao, and Weizhu Chen. 2021.
\newblock \href {https://openreview.net/forum?id=XPZIaotutsD} {{DeBERTa}:
  decoding-enhanced bert with disentangled attention}.
\newblock In \emph{Proc.\ of ICLR}.

\bibitem[{Izsak et~al.(2021)Izsak, Berchansky, and
  Levy}]{izsak-etal-2021-train}
Peter Izsak, Moshe Berchansky, and Omer Levy. 2021.
\newblock \href {https://arxiv.org/abs/2104.07705} {How to train {BERT} with an
  academic budget}.
\newblock In \emph{Proc.\ o EMNLP}.

\bibitem[{Katharopoulos et~al.(2020)Katharopoulos, Vyas, Pappas, and
  Fleuret}]{katharopoulos-et-al-2020}
Angelos Katharopoulos, Apoorv Vyas, Nikolaos Pappas, and Fran\c{c}ois Fleuret.
  2020.
\newblock \href {https://arxiv.org/abs/2006.16236} {Transformers are {RNN}s:
  Fast autoregressive transformers with linear attention}.
\newblock In \emph{Proc.\ of ICML}.

\bibitem[{Keung et~al.(2020)Keung, Lu, Szarvas, and Smith}]{amazon-reviews}
Phillip Keung, Yichao Lu, Gy{\"o}rgy Szarvas, and Noah~A Smith. 2020.
\newblock The multilingual amazon reviews corpus.
\newblock \emph{arXiv preprint arXiv:2010.02573}.

\bibitem[{Lee et~al.(2019)Lee, Cho, and Kang}]{mixout}
Cheolhyoung Lee, Kyunghyun Cho, and Wanmo Kang. 2019.
\newblock Mixout: Effective regularization to finetune large-scale pretrained
  language models.
\newblock \emph{arXiv preprint arXiv:1909.11299}.

\bibitem[{Levesque et~al.(2012)Levesque, Davis, and Morgenstern}]{wnli}
Hector~J. Levesque, Ernest Davis, and Leora Morgenstern. 2012.
\newblock The winograd schema challenge.
\newblock In \emph{Proc.\ of KR}.

\bibitem[{Liu et~al.(2019)Liu, Ott, Goyal, Du, Joshi, Chen, Levy, Lewis,
  Zettlemoyer, and Stoyanov}]{Liu2019RoBERTaAR}
Yinhan Liu, Myle Ott, Naman Goyal, Jingfei Du, Mandar Joshi, Danqi Chen, Omer
  Levy, Mike Lewis, Luke~S. Zettlemoyer, and Veselin Stoyanov. 2019.
\newblock \href {https://arxiv.org/abs/1907.11692} {{RoBERTa}: A robustly
  optimized bert pretraining approach}.

\bibitem[{Maas et~al.(2011)Maas, Daly, Pham, Huang, Ng, and Potts}]{imdb}
Andrew~L. Maas, Raymond~E. Daly, Peter~T. Pham, Dan Huang, Andrew~Y. Ng, and
  Christopher Potts. 2011.
\newblock \href {http://www.aclweb.org/anthology/P11-1015} {Learning word
  vectors for sentiment analysis}.
\newblock In \emph{Proceedings of the 49th Annual Meeting of the Association
  for Computational Linguistics: Human Language Technologies}, pages 142--150,
  Portland, Oregon, USA. Association for Computational Linguistics.

\bibitem[{Mosbach et~al.(2020)Mosbach, Andriushchenko, and
  Klakow}]{glue_instability}
Marius Mosbach, Maksym Andriushchenko, and Dietrich Klakow. 2020.
\newblock On the stability of fine-tuning bert: Misconceptions, explanations,
  and strong baselines.
\newblock \emph{arXiv preprint arXiv:2006.04884}.

\bibitem[{Peng et~al.(2022)Peng, Kasai, Pappas, Yogatama, Wu, Kong, Schwartz,
  and Smith}]{peng-etal-2022-abc}
Hao Peng, Jungo Kasai, Nikolaos Pappas, Dani Yogatama, Zhaofeng Wu, Lingpeng
  Kong, Roy Schwartz, and Noah~A. Smith. 2022.
\newblock \href {https://arxiv.org/abs/2110.02488} {{ABC}: Attention with
  bounded-memory control}.
\newblock In \emph{Proc.\ of ACL}.

\bibitem[{Peng et~al.(2021)Peng, Pappas, Yogatama, Schwartz, Smith, and
  Kong}]{RFA}
Hao Peng, Nikolaos Pappas, Dani Yogatama, Roy Schwartz, Noah~A. Smith, and
  Lingpeng Kong. 2021.
\newblock \href {https://openreview.net/forum?id=QtTKTdVrFBB} {Random feature
  attention}.
\newblock In \emph{Proc.\ of ICLR}.

\bibitem[{Press et~al.(2020)Press, Smith, and Levy}]{press-etal-2020-improving}
Ofir Press, Noah~A. Smith, and Omer Levy. 2020.
\newblock \href {https://arxiv.org/abs/1911.03864} {Improving transformer
  models by reordering their sublayers}.
\newblock In \emph{Proc.\ of ACL}.

\bibitem[{Rajpurkar et~al.(2016)Rajpurkar, Zhang, Lopyrev, and
  Liang}]{rajpurkar-etal-2016-squad}
Pranav Rajpurkar, Jian Zhang, Konstantin Lopyrev, and Percy Liang. 2016.
\newblock \href {https://arxiv.org/abs/1606.05250} {{SQ}u{AD}: 100,000+
  questions for machine comprehension of text}.
\newblock In \emph{Proc.\ of EMNLP}.

\bibitem[{Sharma et~al.(2019)Sharma, Graesser, Nangia, and Evci}]{qqp}
Lakshay Sharma, Laura Graesser, Nikita Nangia, and Utku Evci. 2019.
\newblock \href {http://arxiv.org/abs/1907.01041} {Natural language
  understanding with the quora question pairs dataset}.

\bibitem[{Socher et~al.(2013)Socher, Perelygin, Wu, Chuang, Manning, Ng, and
  Potts}]{sst2-dataset}
Richard Socher, Alex Perelygin, Jean Wu, Jason Chuang, Christopher~D. Manning,
  Andrew~Y. Ng, and Christopher Potts. 2013.
\newblock \href {https://www.aclweb.org/anthology/D13-1170/} {Recursive deep
  models for semantic compositionality over a sentiment treebank}.
\newblock In \emph{Proc.\ of EMNLP}.

\bibitem[{Tjong Kim~Sang and
  De~Meulder(2003)}]{tjong-kim-sang-de-meulder-2003-introduction}
Erik~F. Tjong Kim~Sang and Fien De~Meulder. 2003.
\newblock \href {https://www.aclweb.org/anthology/W03-0419} {Introduction to
  the {C}o{NLL}-2003 shared task: Language-independent named entity
  recognition}.
\newblock In \emph{Proc.\ of CoNLL}.

\bibitem[{Vaswani et~al.(2017)Vaswani, Shazeer, Parmar, Uszkoreit, Jones,
  Gomez, Kaiser, and Polosukhin}]{Vaswani2017AttentionIA}
Ashish Vaswani, Noam Shazeer, Niki Parmar, Jakob Uszkoreit, Llion Jones,
  Aidan~N. Gomez, \L{}ukasz Kaiser, and Illia Polosukhin. 2017.
\newblock \href {https://arxiv.org/abs/1706.03762} {Attention is all you need}.
\newblock In \emph{Proc. of NeurIPS}.

\bibitem[{Wang et~al.(2019)Wang, Pruksachatkun, Nangia, Singh, Michael, Hill,
  Levy, and Bowman}]{superglue}
Alex Wang, Yada Pruksachatkun, Nikita Nangia, Amanpreet Singh, Julian Michael,
  Felix Hill, Omer Levy, and Samuel Bowman. 2019.
\newblock \href {https://arxiv.org/abs/1905.00537} {{SuperGLUE}: A stickier
  benchmark for general-purpose language understanding systems}.
\newblock In \emph{Proc.\ of NeurIPS}.

\bibitem[{Wang et~al.(2018)Wang, Singh, Michael, Hill, Levy, and
  Bowman}]{wang-etal-2018-glue}
Alex Wang, Amanpreet Singh, Julian Michael, Felix Hill, Omer Levy, and Samuel
  Bowman. 2018.
\newblock \href {https://arxiv.org/abs/1804.07461} {{GLUE}: A multi-task
  benchmark and analysis platform for natural language understanding}.
\newblock In \emph{Proc.\ of BlackboxNLP}.

\bibitem[{Wang et~al.(2020)Wang, Li, Khabsa, Fang, and Ma}]{wang2020linformer}
Sinong Wang, Belinda~Z. Li, Madian Khabsa, Han Fang, and Hao Ma. 2020.
\newblock \href {https://arxiv.org/abs/2006.04768} {Linformer: Self-attention
  with linear complexity}.

\bibitem[{Williams et~al.(2018)Williams, Nangia, and Bowman}]{mnli-dataset}
Adina Williams, Nikita Nangia, and Samuel~R. Bowman. 2018.
\newblock \href {https://doi.org/10.18653/v1/n18-1101} {A broad-coverage
  challenge corpus for sentence understanding through inference}.
\newblock In \emph{Proc.\ of NAACL}.

\end{thebibliography}
\bibliographystyle{acl_natbib}

\end{document}